\documentclass{article}
\usepackage{ijcai23}

\usepackage{times}
\usepackage{soul}
\usepackage{url}
\usepackage[hidelinks]{hyperref}
\usepackage[utf8]{inputenc}
\usepackage[small]{caption}
\usepackage{graphicx}
\usepackage{amsmath}
\usepackage{amsthm}
\usepackage{booktabs}
\usepackage{algorithm}
\usepackage{algorithmic}
\usepackage[switch]{lineno}

\usepackage{amsmath}
\usepackage{amssymb}
\usepackage{mathtools}
\usepackage{amsthm}
\usepackage{amsmath}

\DeclareMathOperator*{\argmaxB}{argmax}

\newtheorem{thm}{Theorem}[section]
\newtheorem{lem}{Lemma}[section]

\theoremstyle{definition}

\usepackage[colorinlistoftodos]{todonotes}

\title{pTSE: A Multi-model Ensemble Method for Probabilistic Time Series Forecasting}

\author{
Yunyi Zhou$^1$
\and
Zhixuan Chu$^1$ \thanks{Corresponding author} \and
Yijia Ruan$^1$\and
Ge Jin$^1$\and
Yuchen Huang$^1$\and
Sheng Li$^2$
\affiliations
$^1$Ant Group\\
$^2$University of Virginia\\
\emails
\{zhouyunyi.zyy, yijia.ryj, elvis.jg\}@antgroup.com,
chuzhixuan.czx@alibaba-inc.com,
hyc264276@antfin.com,
shengli@virginia.edu
}
\begin{document}
% \title{\Large pTSE: A Multi-model Ensemble Method for Probabilistic Time Series Forecasting}

\maketitle

\begin{abstract} 
% \small\baselineskip=9pt 
Various probabilistic time series forecasting models have sprung up and shown remarkably good performance. However, the choice of model highly relies on the characteristics of the input time series and the fixed distribution that the model is based on. Due to the fact that the probability distributions cannot be averaged over different models straightforwardly, the current time series model ensemble methods cannot be directly applied to improve the robustness and accuracy of forecasting. 
To address this issue, we propose pTSE, a multi-model distribution ensemble method for probabilistic forecasting based on Hidden Markov Model (HMM). pTSE only takes off-the-shelf outputs from member models without requiring further information about each model. Besides, we provide a complete theoretical analysis of pTSE to prove that the empirical distribution of time series subject to an HMM will converge to the stationary distribution almost surely. Experiments on benchmarks show the superiority of pTSE over all member models and competitive ensemble methods.
\end{abstract}

\section{Introduction}
The common requirements of time series forecasting are not only predicting the expected value of a future target, namely point estimation but also further measuring the uncertainty of the output by predicting its probability distributions, namely probabilistic forecasting. Probabilistic forecasting methods have been extensively studied in the literature, such as deterministic methods that predict the quantiles of the predictive distribution \cite{lim2021temporal}, probabilistic methods that sample future values from a learned approximate distribution \cite{salinas2020deepar,rangapuram2018deep,salinas2019high}), and latent generative models \cite{yuan2019diverse,koochali2021if,rasul2020multivariate}.  
These methods are usually motivated by a particular modeling focus, characterizing certain aspects of the input time series \cite{criteria_classifying_forecasting}. 
For instance, Prophet \cite{Prophet} shows advantages in explicitly characterizing the fundamental time-domain components, i.e., trend and seasonality of time series, while the method in \cite{Temporal_pattern_attention_for_multivariate_time_series_forecasting,chu2023continual,chu2023continuala} digests the frequency domain information for time series to overcome the nonstationary problem. 
On the other hand, the training objective of such probabilistic forecasting methods is usually maximizing a likelihood function that is conventionally assumed to be a fixed distribution, e.g., Gaussian. However, this is not always true for time series. According to \cite{miture_models_timeseries}, real world time series data is more likely to be asymmetric and multi-modal with a mixture of distributions. 
Both aforementioned facts bring challenges for single models.
Therefore, this calls for a new way to integrate the advantages and specificity of diverse models.

To combine the advantages of different models, a popular and competitive solution is model ensemble \cite{ensemble_time_series_forecasting}. 
Generally, there are two main categories of ensemble techniques for time series. 
The first one learns an optimal linear combination of the predicted values returned by each member model \cite{ensemble_approach_ts_demand_forecasting,NN_based_linear_ensemble_time_series} by searching for the optimal weight of each model output. 
The second one trains the member models as weak predictors and then combines them via a boosting-like ensemble step, where member model information, including input features, is typically required for loss reduction \cite{EMD_ensemble,ensemble_based_LSTM,ensemles_of_losclised_models,ensemble_deep_learning_DBN_SVR}. However, to the best of our knowledge, all of the aforementioned studies need to treat all member models as point estimation models, where the ensemble step handles target values not distributions, and thus cannot be directly applied to fulfill probabilistic forecasting model ensemble.
In addition to such practical shortcomings, an adapted theoretical foundation aimed at a probabilistic forecasting ensemble is also highly desired. 

This paper intends to fill these gaps by designing \textit{pTSE} (``probabilistic time series ensemble''), a semi-parametric method, to perform distribution ensemble for probabilistic forecasting of time series. We adopt the idea from the Hidden Markov Model (HMM) to treat the collection of member models as a hidden state space, where the distribution of each observation is determined by its hidden state. 
Then, we incorporate ``mixture quantile estimation'' (MQE) into the classic Baum-Welch algorithm to estimate the distribution of model residuals, which is subsequently used to compute a distribution quantile at the prediction stage. In order to guarantee the generality, we use weighted kernel density estimation (WKDE), a non-parametric method, to approximate the residual distributions, where the sample weight of each residual for a member model is the probability of the corresponding model in the hidden state. A bootstrap method is specifically designed to calculate the optimal bandwidth parameter for each model, which is crucial to the WKDE performance. The whole ensemble step is achieved by inferring the stationary distribution of the HMM, of which the quantile is used for forecasting. We provide a complete theoretical analysis of pTSE to prove that the empirical distribution of time series subject to an HMM will converge to the stationary distribution almost surely, and such a result is non-trivial to guess. \textit{That is to say, the whole time series approximately subjects to the stationary distribution (ensemble distribution) inferred by pTSE within any period of time.} It is worth noting that pTSE takes off-the-shelf model outputs without further requiring implementation details of the member model. This makes pTSE plug-and-play and easily integrated into existing models. 
We evaluate pTSE on synthetic datasets to confirm the theoretical results and then on public data sets to show its superiority over single-model methods as well as ensemble methods designed for point estimation models.

The main contributions of this paper are three folds: (1) We propose pTSE, a multi-model ensemble method for probabilistic forecasting, which only takes off-the-shelf outputs from member models without requiring further information about each model; (2) We theoretically verify the ensemble distribution discovered by our method, which the time series approximately subject to within any period of time; (3) We demonstrate on real-world data sets that our ensemble method produces better performance than all member models as well as competitive point estimation model ensemble methods.

\section{The pTSE Framework}
In this section, we begin with giving a brief overview of HMM as a preliminary in Section \ref{preliminary}.
Then the core idea of pTSE is introduced in Section  \ref{model_foundatio}. 
The distribution evaluation procedure with a mixture quantile estimation method and its parameter selection procedure are described in Section \ref{mqe_section} and Section \ref{boostrap_section}, respectively. Section \ref{mqe_bw_algo} presents the complete procedure of pTSE, including the parameter estimation stage and the prediction stage, where forecasting is made based on the ensemble distribution.

\subsection{Preliminaries}
\label{preliminary}
We first give a brief review of HMM and then introduce its fitting process. 

\textbf{HMM:}
An HMM is a probabilistic model describing the joint probability of a collection of random variables $\{O_1,\ldots,O_T, S_1,\ldots,S_T\}$\cite{gentle_tut_EM}. 
The $O_t$ variables, either continuous or discrete, represent the observations that we can acquire in the real world, while the $S_t$ variables are hidden states corresponding to each $O_t$.
Under an HMM, the random process $\{S_t: t\in \mathbb{N}\}$ is a Markov process satisfying 
$$p(S_{t+1}|S_1,\ldots,S_t)= p(S_{t+1}|S_t).$$
Given the state $S_t$, $O_t$ satisfies
\begin{align}
\label{emission_func}
    p(O_{t}|S_1,\ldots,S_T, O_1, \ldots, O_T)= p(O_t|S_t),
\end{align}
which demonstrates that the distribution of $O_t$ is only determined by the hidden state.

In other words, for continuous variables, Equation \eqref{emission_func} defines the Probability Density Function (PDF) of $O_t$ given $S_t$, and the PDF denoted as 
$f_{k}(o):=p(O_t=o|S_t=k)$, is conventionally termed the ``emission function''.

\textbf{Fitting an HMM:}
Fitting an HMM with a total number of $K$ states to a dataset $\{O_t: t\in \mathbb{N}, 1\leq t\leq T \}$ requires determining the following parameters: (1) transition matrix $\boldsymbol{A}=\left(a_{i,j}\right)_{1\leq i, j\leq K}$ of the underlying Markov process $\{S_t\}$, where $a_{i j}=p(S_{t+1}=j|S_{t}=i)$, (2) a parameter set $\Theta=\{\theta_k\}^K_{k=1}$ of the emission function $f_k(o;\theta_k)$, and (3) initial distribution $\pi=\left(\pi_1,\ldots,\pi_K\right)$, where $\pi_k=p(S_0=k)$.

\begin{figure*}[t]
\centering
\includegraphics[scale=0.35]{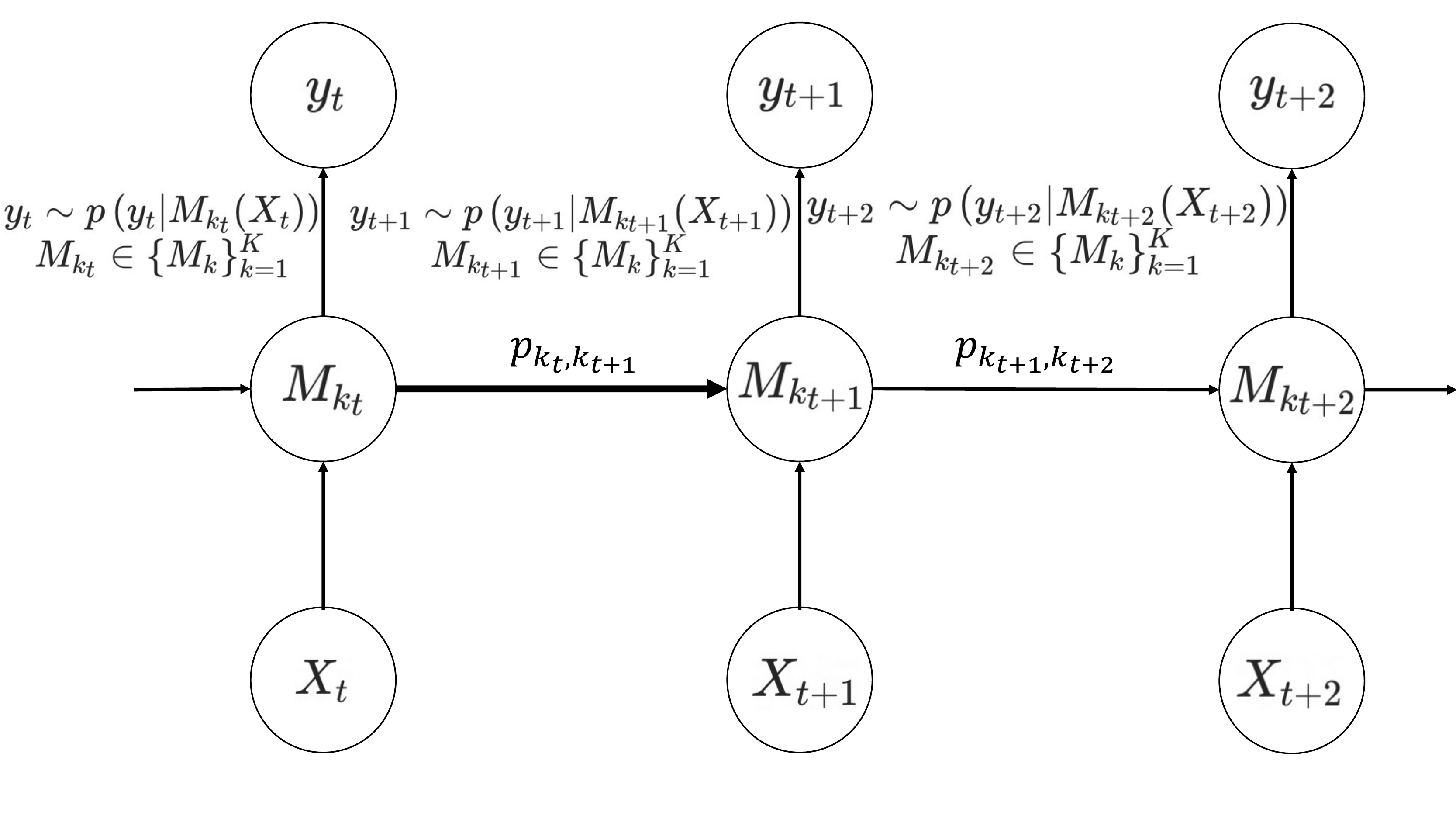}
\vspace{-5mm}
\caption{Markov Property of the Optimal Model Transition Process. At each time $t$, we assume there exists an optimal model $M_{kt} \in \{M_k\}^K_{k=1}$ such that the distribution of $y_t$ is determined by $M_{kt}(X_t)$, where $\{M_k\}^K_{k=1}$ is the set of all $K$ models fitted to the time series $\{y_t\}^T_{t=1}$ and $X_t$ is the feature obtained at $t$ for predicting $y_t$. The optimal model at time $t$ will transfer to a new optimal model $M_{kt+1}$ at time $t+1$ with a probability of $p_{k_t, k_{t+1}}$ (black bold arrow), and we assume this random transition process to be Markov.}
\label{markov_process}
\end{figure*}

The fitting is done by maximum likelihood estimation (MLE) or equivalently 
\begin{align}
\label{mle_HMM}
    \argmaxB_{\boldsymbol{A},\pi,\Theta} p(\{O_t\}^T_{t=1} |\boldsymbol{A},\pi,f_k(O_t;\theta_k\in \Theta)).
\end{align}
However, for a dataset governed by an HMM, the undergoing state process $\{S_t: S_t \in \mathbb{N}, 1\leq S_t \leq K\}$ is unknown in most cases. Therefore, the parameter estimation problem of an HMM is generally solved via an Expectation-Maximization (EM) fashion, which particularly deals with MLE problems with missing data, such as the hidden states. This EM method for fitting HMM is known as the Baum-Welch algorithm \cite{gentle_tut_EM}.

\subsection{Framework Basics}
\label{model_foundatio}
We now introduce the pTSE framework for probabilistic forecasting. A probabilistic forecasting problem usually requires to estimate the conditional distribution of $y_t$, given a trained model $M$ and a feature vector $X_t$, where $X_t$ may contain the history of the time series along with known future information without randomness. In other words, probabilistic forecasting aims to estimate $p(y_t|M(X_t))$.

Suppose a total number of $K$ probabilistic forecasting models, $\{M_k\}^K_{k=1}$, are independently fitted to the same data set $\{ y_t\}^T_{t=1} (T\in \mathbb{N})$. 
We first assume that at each time $t$, there exists an optimal model $M_{k_{t}} \in \{M_k\}^K_{k=1}$ such that the distribution of $y_t$ is determined by the optimal model given the feature $X_t$. Or equivalently, 
$y_t\sim p(y_t|M_{k_t}(X_t))$.

Next, for $y_{t+1}$, we assume that $M_{k_{t}}$ will randomly transfer to a new optimal model $M_{k_{t+1}}$ with a probability of $p_{k_t, k_{t+1}}$ (bold arrow in Figure \ref{markov_process}), and this transition process is a \textbf{Markov process}. We illustrate this idea in Figure \ref{markov_process}.
In order to derive an executable methodology, we denote the PDF $p(y_t|M_{k_t}(X_t))$ as $f^{X_t}_{M_k}(y_t)$, for convenience.

Recall that we intend to use HMM for the model ensemble, we clarify the concepts as follows: the optimal model $M_{k_t}$ corresponds to the hidden state $S_t$, the time series $y_t$ corresponds to the observation $O_t$, and the PDF $f^{X_t}_{M_k}(y_t)$ corresponds to the emission function $f_k(o)$, as introduced in Section \ref{preliminary}. Hence by now, we have established a framework using \textbf{HMM} to capture the relation between probabilistic forecasting models and the target time series.

\begin{figure*}[ht]
\centering
\includegraphics[scale=0.22]{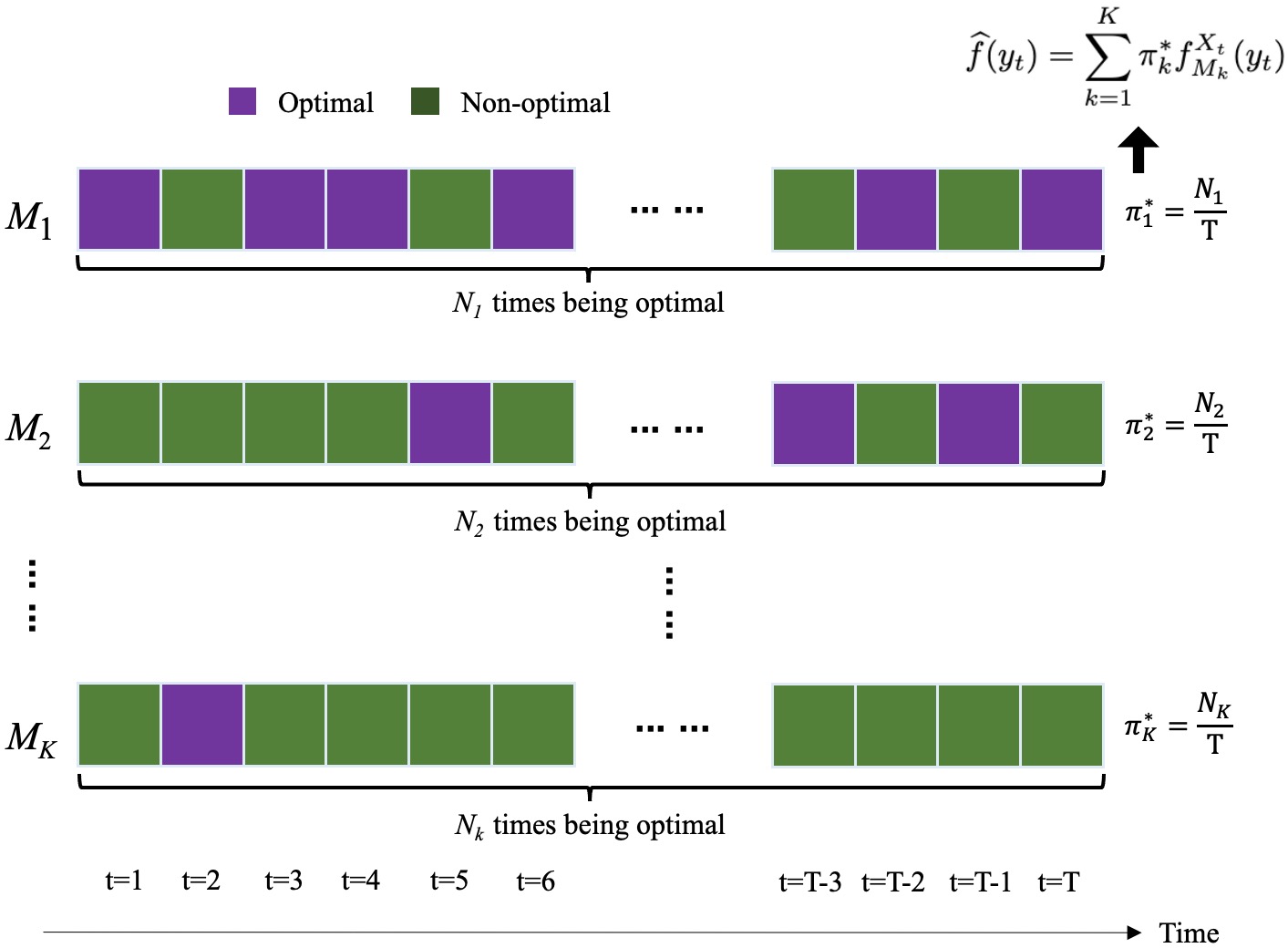}
\caption{Ensemble Distribution. The ensemble PDF is defined as the weighted average of the PDF defined by each member model, where the weights correspond to the average chance of each member model being the optimal model.}
\label{stationary_ensemb}
\end{figure*}

We now introduce the method for evaluating the ensemble weights for each member model $M_k$.
As illustrated by the law of total probability that 
$$ p(y_t)=\sum^K_{k=1} p(y_t|M_k(X_t))p(M_{k}(X_t)),$$
the time series subjects to an ensemble distribution of the component PDF defined by each $M_k(X_t)$ with the weights defined as the probability of each $M_k$ being the optimal model, i.e., $p(M_{k}(X_t))$, where $X_t$ is a known vector without randomness. For simplicity, we estimate $p(M_{k}(X_t))$ as the average chance of $M_k$ being the optimal model within any period of time, denoted as $\pi^*_k$, and the ensemble distribution is estimated as
\begin{align}
    \label{ensemble_form}
    \widehat{f}(y_t)=\sum^K_{k=1}\pi^*_k f^{X_t}_{M_k}(y_t).
\end{align}
We illustrate this idea in Figure \ref{stationary_ensemb}.

Surprisingly, $(\pi^*_1, \ldots, \pi^*_K)$ is nothing but the stationary distribution of the hidden Markov process, which determines the expected frequency of reaching each state. For a Markov process with transition matrix $\boldsymbol{A}$, the stationary distribution is a row vector, $\mathrm{\pi}^*=(\pi^*_1, \ldots, \pi^*_K)$, satisfying $\mathrm{\pi}^* \boldsymbol{A}=\mathrm{\pi}^*$, whose elements are non-negative and sum up to 1. In other words, a key step of our ensemble method is to acquire the stationary distribution of the hidden Markov process, which requires fitting an HMM to the time series $\{y_t\}^T_{t=1}$ given $\{M_k\}^K_{k=1}$ and $\{X_t\}^T_{t=1}$. By substituting the variables of our interest into Equation \eqref{mle_HMM}, the \textbf{ultimate problem} we need to solve is defined as
\begin{align}
\label{mle_ensemble}
    \argmaxB_{\boldsymbol{A},\pi, f^{X_t}_{M_k}} p\left(\{y_t\}^T_{t=1} \bigg |\boldsymbol{A},\pi, f^{X_t}_{M_k}\right).
\end{align}
Although it may seem to be slightly arbitrary to use the stationary distribution to evaluate the ensemble distribution, we have theoretically proved that the time series $\{y_t: t\in \mathbb{N} \}$ approximately subject to this ensemble distribution within any period of time in Section \ref{theoretical_analysis}.

\subsection{Mixture Quantile Estimation}
\label{mqe_section}
Generally, for a probabilistic forecasting method, the PDF $f^{X_t}_{M_k}(y_t)$ is not directly evaluated in the prediction stage; instead, it tends to estimate a quantile of $y_t$, where the $q-$th quantile of $y_t$ is a constant $\tau$ satisfying $p(y_t\leq \tau)=q$.
Therefore, we exploit an MQE framework \cite{mixtures_quantile_regression} to estimate Equation \eqref{ensemble_form}.

The MQE framework first formalizes a $q$-th quantile forecasting model $M$ as
$y=M(X) + \epsilon_q, $
where $X$ is the vector of model inputs, and the error term $\epsilon_q$ is a random variable whose $q-$th quantile is equal to zero. Let $f_{\epsilon_q}(\cdot)$ be the PDF of $\epsilon_q$ and the PDF of $y$ given $M(X)$ is $f_{M}(y)=f_{\epsilon_q}(y-M(X))$. For multiple models, the MQE framework is used to estimate the error term PDF for each model and ensure the $q-$th quantile of $\epsilon_q$ is zero for each model.

Each $y_t$ is then a random variable whose PDF is defined as 
$f^{X_t}_{M_k}(y_t)=f^k_{\epsilon_q}\left(y_t-M_k(X_t)\right)$. The function $f^k_{\epsilon_q}(\cdot)$ is the error term PDF of the member model $M_k$ and can be estimated by the MQE framework. 

% \subsection{Kernel density estimation}
\subsection{Kernel Density Estimation}
\label{boostrap_section}

To perform MQE, we incorporate WKDE, a non-parametric method for PDF estimation. The WKDE is used specifically to estimate the PDF of error term $f^k_{\epsilon_q}(\cdot)$.
A WKDE implies the importance of each sample is unequal, which in our case, is due to the diverse probabilities of different member models being the optimal model for a certain $y_t$. The relation between sample weights and member models will be further discussed in Section \ref{mqe_bw_algo}.

A WKDE obtained from a dataset  $\{y_t\}^T_{t=1}$ is defined as
$\hat{f}(y)=\frac{1}{\sigma\sum^T_{t=1}w_t}\sum^T_{t=1}w_t K\left(\frac{y-y_t}{\sigma}\right), w_t > 0,$
where the bandwidth parameter $\sigma$ essentially influences the performance.
We select the optimal $\sigma$ by a bootstrap method \cite{bootstrap_kde}. For a set of candidate bandwidth parameters $\Sigma=\{\sigma_i\}$, the bootstrap method first chooses an initial value $\sigma_0$. Next, it constructs a total number of $B$ sample sets by resampling from the distribution determined by a WKDE with $\sigma_0$. For each $\sigma_i \in \Sigma$, a WKDE is performed on each resampled sample set, resulting in a PDF $\hat{f}^b_{\sigma_i}(\cdot)$ (corresponding to the $b$th sample set). The selected $\sigma^*$ is the one that minimizes the bootstrap integrated mean squared error (BIMSE), or technically 
\begin{align}
\label{sigma_select}
\sigma^*=\mathrm{argmin}_{\sigma_i \in \Sigma} \frac{1}{B}\sum^B_{b=1}\int\left(\hat{f}^b_{\sigma_i}(o) - \hat{f}_{\sigma_0}(o)\right)^2\mathrm{d}o.
\end{align}
In this work, we adopt the Gaussian kernel for performing WKDE. An estimated PDF $f^k_{\epsilon_q}(\cdot)$ with a selected $\sigma_k$ is denoted as $f^k_{\epsilon_q}(\cdot;\sigma_k)$ for explicitness.

\subsection{The MQE Baum-Welch Algorithm}
\label{mqe_bw_algo}

We now derive the MQE Baum-Welch Algorithm for inferring both $\pi^*$ and $f^{X_t}_{M_k}(y_t)$ in equation \ref{ensemble_form}.

Recall the problem defined in Equation \ref{mle_ensemble} in Section \ref{preliminary}, 
by MQE which refines $f^{X_t}_{M_k}(y_t)$ as $f^k_{\epsilon_q}\left(y_t-M_k(X_t);\sigma_k \right)$, the objective function for the MQE Baum-Welch Algorithm is
\begin{align}
\label{mqe_bw_object}
\argmaxB_{\boldsymbol{A},\pi,\{\sigma_k\}^K_{k=1}} p\left(\{y_t\}^T_{t=1}\bigg|\boldsymbol{A},\pi,  f^k_{\epsilon_q}(\cdot;\sigma_k)\right).
\end{align}

The MQE Baum-Welch Algorithm is still an EM method for resolving Equation \ref{mqe_bw_object}.

During the \textbf{E step}, four quantities $\alpha_k(t)$, $\beta_k(t)$, $\gamma_k(t)$ and $\xi_{i,j}(t)$, need to be prepared, whose definition will be given in the following content. The \textbf{E step} uses the same update equations in the general Baum-Welch Algorithm introduced in \cite{gentle_tut_EM}. Based on the \textbf{E step}, the \textbf{M step} is designed to update the transition matrix $\boldsymbol{A}$, the initial distribution $\pi$, and the PDF $f^k_{\epsilon_q}(\epsilon)$ for each member model, where the MQE method and the bootstrap procedure are performed. The \textbf{EM steps} are repeated until all parameters converge.

The \textbf{stationary distribution} $\pi^*$ in Equation \eqref{ensemble_form} is obtained by setting $\pi^*=\pi$ and repeating $\pi^* = \pi^* \boldsymbol{A}$ until $\pi^*$ converges, where $\pi$ is the estimated initial distribution.
Once $\pi^*$ and $f^k_{\epsilon_q}(\epsilon)$ are determined, the ensemble PDF of $y_{T+h}$ provided with the output of each member model $M_k(X_{T+h})$, i.e. $\hat{f}(y_{T+h};\pi^*, f^k_{\epsilon_q}, M_k(X_{T+h}))$, is defined as
\begin{small}
\begin{align}
\label{ensemble_distribution_result}
\hat{f}(y_{T+h};\pi^*, f^k_{\epsilon_q}, M_k(X_{T+h}))=\sum^K_{k=1}\pi^*_k f^k_{\epsilon_q}(y_{T+h}-M_k(X_{T+h})).
\end{align}
\end{small}

\subsubsection{MQE Baum-Welch Algorithm}
\textbf{\emph{E step}}\\
\noindent\\
\textbf{Updating $\mathbf{\alpha}_k(t)$, $\mathbf{\beta}_k(t)$, $\mathbf{\gamma}_k(t)$ and $\mathbf{\xi}_{i,j}(t)$:} For $ 1\leq i, j\leq K$, define $\alpha_k(t)$ and $\beta_k(t)$ as 

\begin{footnotesize}
\begin{align}
    \label{alph0_update}
    &\alpha_k(1)=\pi_k f^k_{\epsilon_q}(y_1 - M_k(X_1)),\, k=1,\ldots,K,\\
    \label{alpha_update}
    &\alpha_j(t+1)=\left(\sum^K_{i=1}\alpha_i(t)a_{ij}\right) f^j_{\epsilon_q}(y_{t+1} - M_j(X_{t+1});\sigma_k),%\, 1\leq i, j\leq K,
\end{align}
\end{footnotesize}

and
\begin{footnotesize}
\begin{align}
    \label{beta0_update}
    \beta_k(T)&=1,\, k=1,\ldots,K,\\
    \label{beta_update}
    \beta_j(t)&=\left(\sum^K_{i=1}a_{ij} f^j_{\epsilon_q}(y_{t+1} - M_j(X_{t+1});\sigma_k)\right) \beta_j(t+1),%\, 1\leq i, j\leq K,
\end{align}
\end{footnotesize}
where $\alpha_k(t),\,\beta_k(t)$ are termed forward probability and backward probability respectively.

Based on $\alpha_k(t)$ and $\beta_k(t)$, $\gamma_k(t)$ and $\xi_{i,j}(t)$ are defined as:

\begin{align}
    \label{gamma_update}
    \gamma_k(t)&=\frac{\alpha_k(t)\beta_k(t)}{\sum_j\alpha_j(t)\beta_j(t)}),
\end{align}
which is the probability of being in state $k$ at time $t$, and 

\begin{align}
    \label{xi_update}
    \xi_{i, j}(t)&=\frac{\gamma_i(t)a_{i,j} f^j_{\epsilon_q}(y_{t+1} - M_j(X_{t+1});\sigma_k) \beta_j(t+1)}{\beta_i(t)},
\end{align}

which is the probability of being in state $i$ at time $t$ and in state $j$ at time $t+1$.

\noindent \textbf{\emph{M step}}

\textbf{Updating $\boldsymbol{A}$ and $\boldsymbol{\pi}$:} Each element $a_{ij}$ of the transition matrix $\boldsymbol{A}$ is estimated as the expected number of transitions from state $i$ to state $j$ relative to the expected total number of transitions away from $i$ \cite{gentle_tut_EM}, which is
\begin{align}
\label{transition_update}
a_{i,j}=\frac{\sum^{T_1}_{t=1}\xi_{i,j}(t)}{\sum^{T_1}_{t=1}\gamma_i(t)}.
\end{align}
For the initial distribution $\pi=(\pi_1,\ldots,\pi_K)$, the $k$-th element is estimated as the expected relative frequency in the $k$-th state at time 1, or
\begin{align}
\label{pi_update}
\pi_k=\gamma_k(1).
\end{align}
In summary, Equations \eqref{alph0_update}-\eqref{pi_update} aim at updating the transition matrix $\boldsymbol{A}$ and the initial distribution $\pi$, with $f^k_{\epsilon_q}(\epsilon)$ fixed.

\textbf{Selecting $\mathbf{\sigma_k}$ for $\mathbf{f^k_{\epsilon_q}(\epsilon)}$:} Our MQE method subsequently updates the emission function or the PDF $f^k_{\epsilon_q}(\epsilon)$. It first selects a bandwidth parameter $\sigma_k$ for each $f^k_{\epsilon_q}(\epsilon)$ via the bootstrap method. In the bootstrap procedure, $\gamma_k(t)$ is used as the sample weight of each $\epsilon^t_k$, where $\epsilon^t_k=y_t-M_k(X_t)$.
We use $\gamma_k(t)$ as sample weight because it is the probability of $M_k$ being the hidden state of $y_t$ \cite{gentle_tut_EM}. Therefore, for each member model, samples with larger probability should be paid more attention to.

\textbf{Updating $\mathbf{f^k_{\epsilon_q}(\epsilon)}$ by MQE:} Based on $\sigma_k$, a simple weighted-KDE-like update equation for $f^k_{\epsilon_q}(\epsilon)$ would be $\sum^T_{t=1} \frac{\gamma_k(t)}{\sigma_k \sum_t\gamma_k(t)}K\left(\frac{\epsilon-\epsilon^t_k}{\sigma_k }\right)$. However, this update equation ignores which quantile is being estimated by the member models. To make the method focus on the $q$-th quantile, $f^k_{e_q}(\epsilon)$ is updated as 
$$f^k_{\epsilon_q}(\epsilon)=\sum^T_{t=1}\sum^2_{l=1}\mathbf{I}^k_{t,l} W^k_l\gamma_k(t) \frac{1}{\sigma_k}K\left(\frac{\epsilon-\epsilon^t_k}{\sigma_k}\right), $$
where $\mathbf{I}^k_{t,1}=\mathbb{I}_{\{\epsilon^t_k \leq 0\}}$\footnote{The indicator function $\mathbb{I}_{\{\cdot\}}$ is a 0/1 valued function, which is defined as $\mathbb{I}_{\{c\}}=1$ if $c$ is true, and 0 otherwise.} and $\mathbf{I}^k_{t,2}=\mathbb{I}_{\{\epsilon^t_k > 0\}}$ \cite{mixtures_quantile_regression}.  The constants $W^k_1$ and $W^k_2$ constrain $f^k_{\epsilon_q}(\epsilon)$ to have 0 as the $q-$th quantile, while normalizing the integral of $f^k_{\epsilon_q}(\epsilon)$ equal to 1. By defining $v_{k,t}=\int^0_{-\infty} \frac{1}{\sigma_k} K\left(\frac{\epsilon - \epsilon^t_{k}}{\sigma_k} \right)\mathrm{d}\epsilon$,  $W^k_1$ and $W^k_2$ are acquired by solving the following linear equation systems,
\begin{align*}
\left  \{
        \begin{aligned}
        &\sum^T_{t=1}\sum^2_{l=1} \mathbf{I}^k_{t,l}  W^k_l\gamma_k(t)=1,  \\
        &\sum^T_{t=1}\sum^2_{l=1} \mathbf{I}^k_{t,l} v_{k,t}  W^k_l\gamma_k(t)=q.
        \end{aligned} 
        \right.
\end{align*}

% \subsubsection{Prediction step}
\subsubsection{Prediction Step}
\noindent
Once the learning procedure stops, the $q-$th quantile $\tau$ at time $T+h$ is obtained by solving Equation \eqref{ensemble_quantile}
\begin{align}
\label{ensemble_quantile}
    \int^\tau_{-\infty} \sum_k \pi^*_{k} f^k_{\epsilon_q}\left(y-M_k(X_{T+h})\right) \mathrm{d}y=q
\end{align}

Therefore, the future $q$-th quantile for a time series is estimated from ensemble PDF with Equation \eqref{ensemble_quantile} and can be directly used at the prediction stage.

\section{Theoretical Analysis}
\label{theoretical_analysis}
In this section, we present the theoretical results of our work. The core idea is to evaluate the limit of the empirical distribution of a sample set originated by an HMM. The results are non-trivial because a random process $\{O_t\}^{+\infty}_{t=0}$ generated from an HMM is not necessarily a Markov process. Strictly speaking, Equation \eqref{condition_all} is not always equal to Equation \eqref{condition_t_1} for a given initial distribution $\pi$, where $\mathbf{F}(o)=\left(f_1(o),\ldots,f_K(o)\right)^T$ and $f_k(o)$ is the emission function as introduced in \ref{preliminary}.
\begin{align}
\label{condition_all}
&p(O_{t+1} \leq \tau_{t+1}|O_0\leq \tau_0,\ldots, O_t\leq \tau_t) 
 \\\notag
&=\frac{\pi \boldsymbol{A}\int^{\tau_0}_{-\infty}\mathrm{diag}\left(\mathbf{F}(o)\right)\mathrm{d}o\cdots \boldsymbol{A}\int^{\tau_{t+1}}_{-\infty}\mathrm{diag}\left(\mathbf{F}(o)\right)\mathrm{d}o\mathbf{1}}{\pi \boldsymbol{A}\int^{\tau_0}_{-\infty}\mathrm{diag}\left(\mathbf{F}(o)\right)\mathrm{d}o\cdots \boldsymbol{A}\int^{\tau_{t}}_{-\infty}\mathrm{diag}\left(\mathbf{F}(o)\right)\mathrm{d}o \boldsymbol{A}\mathbf{1}}.
\end{align}
\begin{align}
\label{condition_t_1}
&p(O_{t+1} \leq \tau_{t+1}|O_t\leq \tau_t)
\\ \notag&=
\frac{\pi \boldsymbol{A}^{t-1}\int^{\tau_{t}}_{-\infty}\mathrm{diag}\left(\mathbf{F}(o)\right)\mathrm{d}o \boldsymbol{A} \int^{\tau_{t+1}}_{-\infty}\mathrm{diag}\left(\mathbf{F}(o)\right)\mathrm{d}o\mathbf{1}}{\pi \boldsymbol{A}^{t-1}\int^{\tau_{t}}_{-\infty}\mathrm{diag}\left(\mathbf{F}(o)\right)\mathrm{d}o \boldsymbol{A}\mathbf{1}}.
\end{align}

We summarize our theoretical work in four lemmas and one theorem. Lemma \ref{right_eigen} is a straightforward statement of the property of a transition matrix. Both Lemma \ref{limit} and Lemma \ref{BC} illustrate convergence. Lemma \ref{limit} computes the final limit with the help of Lemma \ref{right_eigen}. Lemma \ref{BC} assures the existence of the limit of the sum of a random variable sequence, providing fast dependence decay. Lemma \ref{cov_rate}, indicating the exponential convergence rate of a Markov Process, turns out to be a guarantee for the condition in Lemma \ref{BC}. 
Remarkably, based on Lemma \ref{right_eigen}-\ref{cov_rate},
Theorem \ref{emp_limit} indicates
that the empirical distribution of a dataset sampled from an HMM converges to $\sum^K_{k=1}\pi^*_k f_k(o)$ almost surely. In other words, these HMM samples are approximately subject to an ensemble distribution of the emission functions, $f_k(o)$, with the weights determined by the stationary distribution regardless of the time window. In the case of pTSE, each emission function $f_k(y_t)$ is defined as $f^k_{\epsilon_q}(y_t-M_k(X_t))$. Hence as a corollary, within any period of time, the time series $\{y_t: t\in \mathbb{N}\}$ approximately subjects to
the ensemble distribution in Equation \eqref{ensemble_distribution_result}, estimated by the MQE Baum-Welch Algorithm.

\begin{figure*}[th!]
\centering
\includegraphics[scale=0.26]{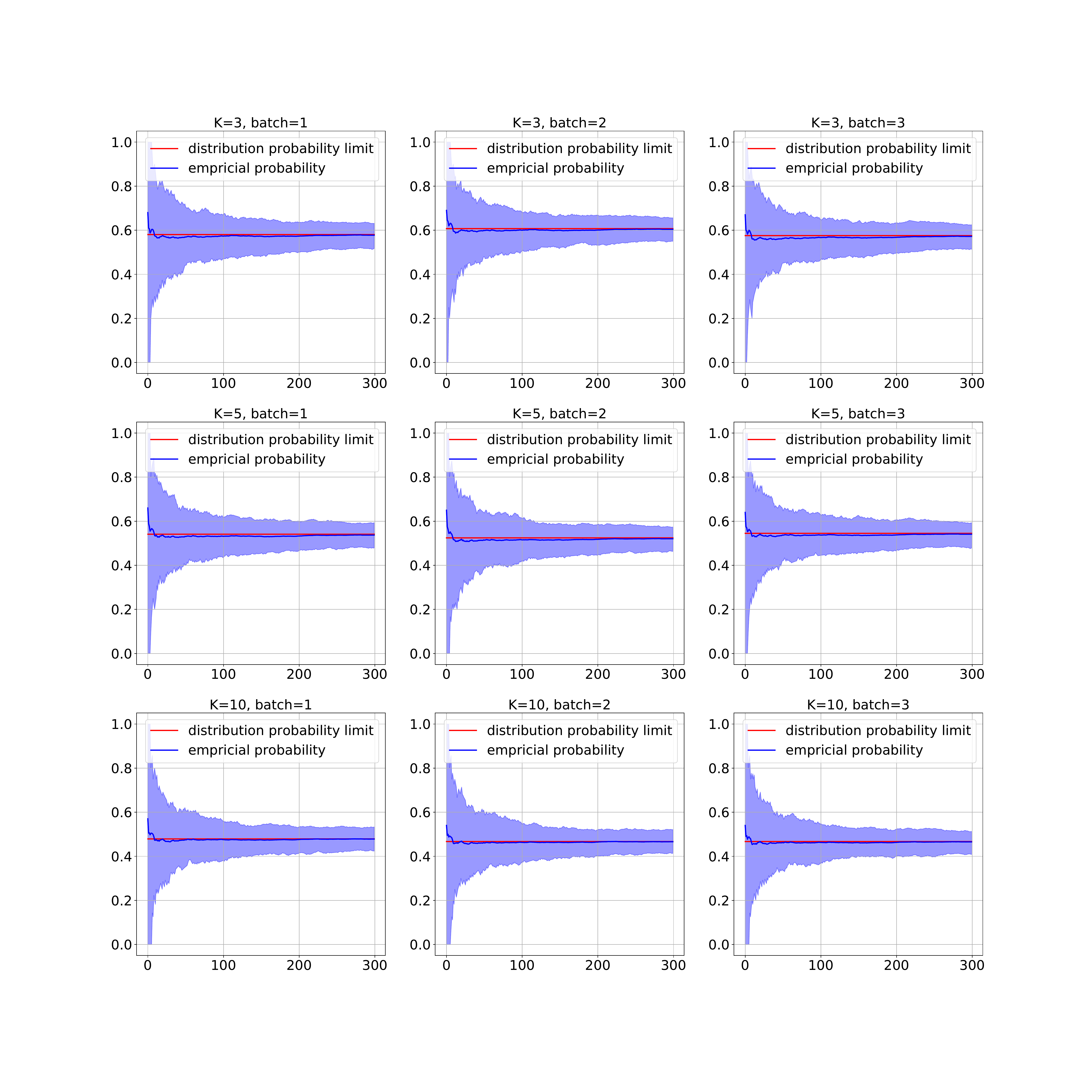}
\vspace{-50pt}
\caption{Average empirical probability (blue line) across 100 simulations vs. theoretical limit (red line) when $\tau=0.5$. The blue shadow represents the $95\%$ confidence interval. Three batches of a 100-time-simulation dataset are presented for each $K$.}
\label{fig_mc_simulation}
\end{figure*}

\begin{lem}
\label{right_eigen}
Let $A$ be the transition probability matrix of an HMM, and let
$$\mathbf{1} = \left(1, 1, \ldots,1\right)^{\mathrm{T}},$$ 
then $\mathbf{1}$ is a right eigenvector of eigenvalue 1 of $A$, i.e.
$$\boldsymbol{A} \mathbf{1} = \mathbf{1}$$
\end{lem}
The proof is straightforward.

\begin{lem}
\label{limit}
{Let $A$ be the transition probability matrix of an HMM, $\pi^*$ be the stationary distribution, and $\pi$ be any initial distribution, let $\mathbf{F}(o)$ be the vector-valued emission function of all states, then 
\begin{small}
\begin{align}
\label{expect_limit}
&\lim_{T\rightarrow+\infty}\frac{1}{T}\sum^{T-1}_{t=0}\pi \boldsymbol{A}^{t}\int^\tau_{-\infty}\mathrm{diag}\left(\mathbf{F}(o)\right)\mathrm{d}o \boldsymbol{A}^{T-1-t}\mathbf{1} =\pi^*\int^\tau_{-\infty}\mathbf{F}(o)\mathrm{d}o&
\end{align}
\end{small}
}
\end{lem}

\begin{lem}
\label{BC}

{
Let $\{E_n: n \geq 1\}$ be a sequence of events and $S_n=\sum^n_{k=1}\mathbb{I}_{E_k}$, 
if there exists $\{\rho_n\}^{\infty}_{n=1}$ satisfying $\sum_n |\rho_n| \leq+\infty$, % \in l^1 $ 
such that, for any $i\neq j$,

$$p(E_i \cap E_j)-p(E_i)p(E_j)\leq \rho_{|i-j|} \sqrt{p(E_i)p(E_j)},$$
and if $\lim_{n\rightarrow +\infty} \mathbb{E}(S_n) = +\infty $,then,
$$\lim_{n\rightarrow +\infty}\frac{S_n}{\mathbb{E}(S_n)}=1, a.s.$$
}
\end{lem}

\begin{table*}[th!]
\centering
% \caption{Table A presents the experiment results of our model pTSE and its member models, and Table B provides the comparison with another two ensemble models on three real datasets. The best results are marked in bold (lower is better).}
% \label{real_data_results}

\label{real_data_results_a}
\begin{tabular}{llllllllll}
\hline
           (\textbf{A}) & \multicolumn{3}{c}{Traffic}                      & \multicolumn{3}{c}{Electric}                     & \multicolumn{3}{c}{Solar Energy}                 \\ \hline
Model       & 0.5-risk       & 0.9-risk       & AVG            & 0.5-risk       & 0.9-risk       & AVG            & 0.5-risk       & 0.9-risk       & AVG            \\ \hline
DeepAR      & 0.151          & 0.302          & 0.227          & 0.083          & 0.070          & 0.076          & \textbf{0.435} & 0.253          & 0.344          \\
SFF         & 0.235          & 0.471          & 0.353          & 0.084          & 0.047          & 0.066          & 0.509          & 0.278          & 0.393          \\
TFT         & 0.184          & 0.367          & 0.275          & 0.118          & 0.062          & 0.090          & 0.487          & \textbf{0.184} & 0.336          \\
Transformer & 0.163          & 0.326          & 0.244          & 0.087          & 0.058          & 0.072          & 0.499          & 0.270          & 0.385          \\ \hline
pTSE (Ours)  & \textbf{0.150} & \textbf{0.106} & \textbf{0.128} & \textbf{0.079} & \textbf{0.042} & \textbf{0.061} & 0.451          & 0.210          & \textbf{0.331} \\ \hline
\end{tabular}

\label{real_data_results_b}
\begin{tabular}{llllllllll}
\hline
         (\textbf{B})  & \multicolumn{3}{c}{Traffic}                      & \multicolumn{3}{c}{Electric}               & \multicolumn{3}{c}{Solar Energy}                 \\ \hline
Model      & 0.5-risk       & 0.9-risk       & AVG            & 0.5-risk & 0.9-risk       & AVG            & 0.5-risk       & 0.9-risk       & AVG            \\ \hline
ModelRank  & \textbf{0.150} & 0.108          & 0.129          & \textbf{0.073}    & 0.049          & \textbf{0.061} & 0.457          & \textbf{0.210} & 0.334          \\
FFORMA     & 0.165          & 0.112          & 0.139          & 0.081    & 0.070          & 0.076          & 0.459          & 0.284          & 0.372          \\ \hline
pTSE (Ours) & \textbf{0.150} & \textbf{0.106} & \textbf{0.128} & 0.079    & \textbf{0.042} & \textbf{0.061} & \textbf{0.451} & \textbf{0.210} & \textbf{0.331} \\ \hline
\end{tabular}

\caption{
The experimental results of 
our model (pTSE), member models, and another two ensemble models on three real datasets. The best results are marked in bold (lower is better).}
\label{real_data_results}
\end{table*}

\begin{lem}
\label{cov_rate}
{Let $A$ be the transition probability matrix of a Markov Process with non-zero elements and $\pi^*$ be the stationary distribution. Then, there exists a constant $C$, such that the following inequality holds,
$$||\boldsymbol{A}^t-\mathbf{1}\pi^*||_F\leq Ct^{J-1}\lambda^{t-J+1}_*,$$
where $J$ is the size of the largest Jordan block of $\boldsymbol{A}$, and $\lambda^{t-J+1}_*$ is the largest absolute value of the eigenvalues smaller than $1$ of $\boldsymbol{A}$. $||\cdot||_F$ is the Frobenius Norm.
}
\end{lem}

\begin{thm}
\label{emp_limit}
Let $\left\{ O_t\right\}^T_{t=1} $ be the observations generated from an HMM whose transition matrix has non-zero elements, $\pi^*$ be the stationary distribution,
 and $\mathbf{F}(o)$ be the vector-valued emission function of all states. Define $\hat{F}_T(\tau)$, the empirical distribution of  $\left\{ O_t\right\}^T_{t=1} $, as 
$$\hat{F}_T(\tau) = \frac{1}{T}\sum^T_{t=1}\mathbb{I}_{\{O_t<=\tau\}} $$
then,
$$\lim_{T\rightarrow+\infty}\hat{F}_T(\tau) = \pi^*\int^\tau_{-\infty}\mathbf{F}(o)\mathrm{d}o,\,a.s.$$
\end{thm}

\section{Numerical Experiments}
In this section, we first present a few synthetic data experiments to verify our theoretical results. Next, we apply the pTSE to publicly available datasets to test the performance.

\subsection{Synthetic Data Analysis}
We simulated random sequences governed by HMM structures. We set $K=3,\,5,\, 10$ and $T=1000$. The transition matrix $\boldsymbol{A}$ is chosen by first generating a matrix of uniformly distributed random numbers and then normalizing the matrix to ensure the sum of elements of each row equals $1$. The emission function $f_k(o)$ for each state $k$ is simply set to a Gaussian distribution as $\mathcal{N}( 0.2k, \sqrt{k}+1),\,\left(k=1, \ldots, K\right)$. For each set of $\{K, T, \boldsymbol{A}, \mathbf{F}(o)\}$, we run the simulation procedure for $100$ times, where during each time, an initial distribution $\pi^0$ is randomly chosen. The results are presented in Figure \ref{fig_mc_simulation}. The empirical probability, $\hat{F}(\tau)$, shows fast convergence to $\pi^*\int^{\tau}_{-\infty}\mathbf{F}(o)\mathrm{d}o$, after $T=50$ for all simulated datasets, regardless of the state number $K$ or the transition matrix $\boldsymbol{A}$.

\subsection{Real World Data Analysis}
We evaluate the performance of pTSE on three challenging real-world benchmark datasets, i.e., solar energy \footnote{https://www.nrel.gov/grid/solar-power-data.html}, electricity, traffic \cite{temp_regu_maxtrix_factorization_time_series}.%, NN5, and Tourism.
The solar energy, electricity, and traffic datasets contain hourly measurements from 137, 370, and 963 time series, respectively.
For these three datasets, each model would perform an iterative prediction task of forecasting the future values for a 24-hour horizon after being trained on the past 168-hour (past week) data. The model performance would be evaluated on a 7-day-horizon test set.

Four of the most popular probabilistic forecasting models are selected as the member models: SimpleFeedForwardEstimator (SFF), Transformer, DeepAR, and TemporalFusionTransformer (TFT). As in \cite{deepAR}, we use $q$-risk metrics (quantile loss) to quantify the accuracy of a $q$-th quantile of the predictive distribution.
Table \ref{real_data_results} presents 0.5-risk, 0.9-risk, and the average risk of the output corresponding to each method. We also present a comparison of pTSE with existing time series model ensemble methods, FFORMA \cite{FFORMA}, a meta-learning approach, and a model ranking-based ensemble method \cite{rankingbase}. Hyperparameters of the two ensemble methods are set up as recommended in the original papers. Results in Table \ref{real_data_results} show that pTSE outperforms all member models on average across all datasets. The performance of pTSE is relatively more stable in the electricity and traffic datasets, as it ranked highest in all three competitions in each case. On the solar energy dataset, pTSE maintained the best performance in the average loss. Compared with the other two ensemble methods, pTSE shows a significant advantage on three datasets.

\section{Conclusion}
\label{conclusion}
We present pTSE, a semi-parametric multi-model ensemble methodology for probabilistic forecasting. Our method takes off-the-shelf model outputs without requiring further information. We theoretically prove the empirical distribution of time series subject to an HMM will converge to the stationary distribution almost surely. We use the synthetic data to verify the validity of our theory and conduct extensive experiments on three benchmark datasets to demonstrate the superiority of pTSE over each member model and other model ensemble methods. Nevertheless, it should be pointed out that the improvement by pTSE or any other ensemble methods is essentially limited by the performance of member models.

\bibliographystyle{named}
\bibliography{ijcai23}

\end{document}